\documentclass{sig-alternate-2013}
\usepackage{times}
\usepackage{latexsym}
\usepackage{url}
\usepackage{graphicx}
\usepackage{tabulary}
\usepackage{amsmath}
\usepackage{epstopdf}
\usepackage{array,paralist}
\newfont{\mycrnotice}{ptmr8t at 7pt}
\newfont{\myconfname}{ptmri8t at 7pt}
\let\crnotice\mycrnotice%
\let\confname\myconfname%

\permission{Permission to make digital or hard copies of all or part of this work for personal or classroom use is granted without fee provided that copies are not made or distributed for profit or commercial advantage and that copies bear this notice and the full citation on the first page. Copyrights for components of this work owned by others than ACM must be honored. Abstracting with credit is permitted. To copy otherwise, or republish, to post on servers or to redistribute to lists, requires prior specific permission and/or a fee. Request permissions from Permissions@acm.org.}
\conferenceinfo{DocEng'15,}{September 8-11, 2015, Lausanne, Switzerland.} 
\copyrightetc{\copyright~2015 ACM. ISBN \the\acmcopyr}
\crdata{978-1-4503-3307-8/15/09\ ...\$15.00.\\
DOI: http://dx.doi.org/10.1145/2682571.2797061}

\newcommand{\mytilde}{\raise.17ex\hbox{$\scriptstyle\mathtt{\sim}$}}

\let\oldbibliography\thebibliography
\renewcommand{\thebibliography}[1]{\oldbibliography{#1}
\setlength{\itemsep}{-1pt}} 

\numberofauthors{3} 
\author{
\alignauthor
Siddhartha Banerjee\\
       \scriptsize{\affaddr{The Pennsylvania State University}\\
			 \affaddr{College of IST}\\
       \affaddr{PA, USA}\\
       \email{sub253@ist.psu.edu}}
\alignauthor
Prasenjit Mitra \\
\scriptsize{\affaddr{Qatar Computing Research Institute}\\
       \affaddr{Hamad Bin Khalifa University}\\
       \affaddr{Doha, Qatar}\\
       \email{pmitra@qf.org.qa}}
\alignauthor 
Kazunari Sugiyama\\
\scriptsize{\affaddr{National University of Singapore}\\
\affaddr{School of Computing}\\
\affaddr{Singapore\\
 \email{sugiyama@comp.nus.edu.sg}}
}}
\title{Generating Abstractive Summaries \\ 
from Meeting Transcripts}

\begin{document}
\maketitle

\begin{abstract}
Summaries of meetings are very important as they convey the essential content of discussions in a concise form. 
Both participants and non-participants are interested in the summaries of meetings to plan for their future work. 
Generally, it is time consuming to read and understand the whole documents. Therefore, summaries play an important role 
as the readers are interested in only the important context of discussions. 
In this work, we address the task of meeting document summarization. 
Automatic summarization systems on meeting conversations developed so far have been primarily extractive, 
resulting in unacceptable summaries that are hard to read. 
The extracted utterances contain disfluencies that affect the quality of the extractive summaries. 
To make summaries much more readable, we propose an approach to generating abstractive summaries by fusing important content from several utterances. 
We first separate meeting transcripts into various topic segments, and then identify the important utterances in each segment using a supervised learning approach. 
The important utterances are then combined together to generate a one-sentence summary. 
In the text generation step, the dependency parses of the utterances in each segment are combined together to create a directed graph. The most informative and well-formed sub-graph obtained by integer linear programming (ILP) is selected to generate a one-sentence summary for each topic segment. The ILP formulation reduces disfluencies by leveraging grammatical relations that are more prominent in non-conversational style of text, and therefore generates summaries that is comparable to human-written abstractive summaries. Experimental results show that our method can generate more informative summaries than the baselines. In addition, readability assessments by human judges as well as log-likelihood estimates obtained from the dependency parser show that our generated summaries are significantly readable and well-formed.
\end{abstract}

\newpage


\section{Introduction}
\label{sec:Intro}
Meeting summarization helps both participants and non-participants by providing a short and concise snapshot of the most important content discussed in the meetings. While previous work on meeting summarization was primarily extractive~\cite{garg2009clusterrank,gillick2009global}, a recent study showed that people generally prefer abstractive summaries~\cite{murray2010generating}. 
\begin{figure*}[t]
	\centering
		\fbox{\includegraphics[width =0.95\textwidth, height=8.8cm]{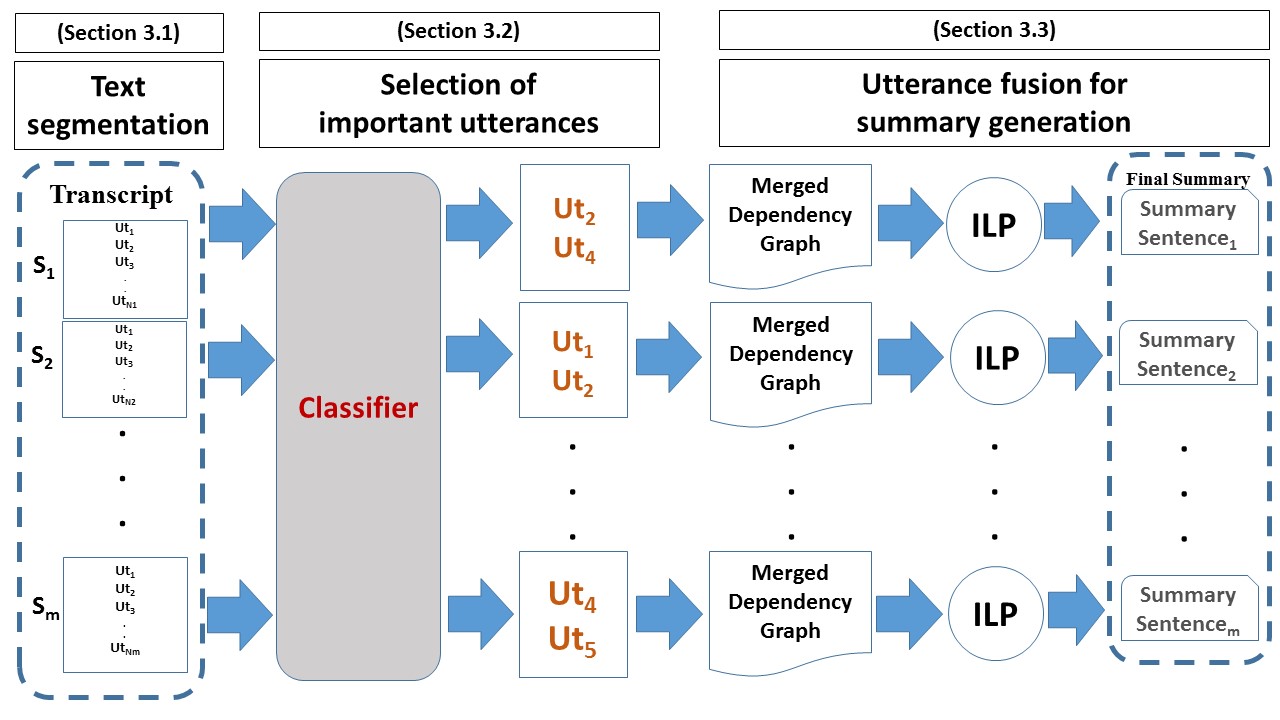}}
	\caption{Our meeting summarization system overview.}
	\label{fig:schematic}
\end{figure*}

\begin{table}[t]
\centering
\caption{Two sets of extractive summaries along with the corresponding gold standard human generated abstractive summaries from a meeting in the AMI corpus~\cite{carletta2006ami}. 
Set~2 follows Set~1 in the actual meeting transcript. 
``A,'' ``B'' and ``D'' refer to three distinct speakers in the meeting.}
\begin{tabular}{|>{\everypar{\hangindent 0.4cm}}p{8cm}|}
\hline
\textbf{Set 1:} \textbf{Human-generated extractive summary}\\
\hline
\textbf{D:} um as well as uh characters.\\
\textbf{D:} um different uh keypad styles and s symbols.\\
\textbf{D:} Well right away I'm wondering if there's um th th uh, like with DVD players, if there are zones.\\
\textbf{A:} Cause you have more complicated characters like European languages, then you need more buttons. \\
\textbf{D:} I'm thinking the price might might appeal to a certain market in one region, whereas in another it'll be different, so\\ 
\textbf{D:} kay trendy probably means something other than just basic\\ 
\hline 
\end{tabular}
\begin{tabular}{|p{8cm}|}
\noindent{\textbf{Abstractive summary:} The team then discussed various features to consider in making the remote.} \\
\hline
\end{tabular}
\begin{tabular}{|>{\everypar{\hangindent 0.4cm}}p{8cm}|}
\textbf{Set 2:} \textbf{Human-generated extractive summary}\\
\hline
\textbf{B:} Like how much does, you know, a remote control cost.\\ 
\textbf{B:} Well twenty five Euro, I mean that's um that's about like eighteen pounds or something.\\
\textbf{D:} This is this gonna to be like the premium product kinda thing or \\
\textbf{B:} So I don't know how how good a remote control that would get you. Um.\\ 
\hline
\end{tabular}
\begin{tabular}{|p{8cm}|}
\textbf{Abstractive summary:} The project manager talked about the project finances and selling prices. \\
\hline
\end{tabular}
	\label{tab:MeetingAbsExample}
\end{table}

Table~\ref{tab:MeetingAbsExample} shows the human-written abstractive summaries along with the human-generated extractive summaries 
from the AMI corpus~\cite{carletta2006ami}. 
Set~1 and Set~2 show two different topics discussed in the meeting -- \textit{design features} and \textit{finances}.
We have skipped other intervening utterances not included in the extractive summary. 
%
As shown in Table~\ref{tab:MeetingAbsExample}, 
the utterances are highly noisy and contain unnecessary information. 
Even if an extractive summarizer can accurately classify these utterances as ``important'' 
and generate a summary, it is usually hard to read and synthesize information from such summaries. 
In contrast, the human written summaries are compact and written in non-conversational style. They are more readable than the extractive summaries and preserve the most important information. 

Previous approaches to abstractive meeting summarization have relied on template-based~\cite{wang2013focused} or word-graph fusion-based~\cite{mehdad2013abstractive} methods. 
The template-based method was applied to the generation of 
\textit{focused summaries}.\footnote{Focused summary refers to summaries on specific aspects of the meeting such as actions, decisions, etc.} 
Template-based generation is feasible in the case where the type of the summary is known apriori; 
however, our work does not make any assumptions on the type of the summary to be generated. 
The word-graph fusion-based technique, on the contrary, used an unsupervised approach to fuse a cluster of utterances generated using an entailment graph-based approach. 
However, this method did not take into consideration any grammatical dependencies between the words, 
resulting in ungrammatical output in several cases. 

In this work, we propose an automatic way of generating short and concise abstractive summaries of meetings. 
Every meeting is usually comprised of several sub-topics~\cite{hsueh2006automatic}.
As shown in Table~\ref{tab:MeetingAbsExample}, the participants discuss different aspects 
in Set~1 and Set~2. 
A well-formed abstractive summary should identify the most important aspects discussed in the meeting. 
In other words, if we can summarize the important information from every aspect, we can generate an informative summary that highlights the salient elements of the meeting. 
Therefore, we need to determine the boundaries where significant topic changes happen to isolate different aspects. Further, to generate a summary for each segment (topic), we should be able to fuse information from multiple utterances on that topic and retain the most informative constituents. Simultaneously, we should also generate grammatical output to ensure that the final summaries are well-formed and readable. 

Figure~\ref{fig:schematic} shows the overview of our proposed meeting summarization system. 
As shown in Figure~\ref{fig:schematic}, initially, a meeting transcript is divided into several topic segments $S_{i}$ ($i=1,2,\ldots ,m$), 
where each segment contains $N_{i}$ utterances ($Ut_{N_{1}},\ldots ,Ut_{N_{i}}$) on a specific topic. Previous work on meeting summarization~\cite{murray2008summarizing} has shown that \textit{lexical cohesion} is an important indicator in topic identification in meetings. We experiment with two different lexical-cohesion based text segmentation algorithms: LCSeg~\cite{galley2003discourse} and unsupervised Bayesian topic segmentation~\cite{eisenstein2008bayesian}. Only a few utterances contain information that is worthy of being included in the summary. 
Therefore, we introduce an extractive summarization component. 
To identify the most important (summary-worthy) utterances, 
we employ a supervised learning approach to construct a classifier by using content and discourse-level features. 
We parse the important utterances in each segment using a dependency parser, and then fuse the corresponding dependency graphs together to form a directed graph (\textit{merged dependency graph}). The directed graph consists of the words in the utterances as the nodes, while the edges represent the grammatical relations between the words. 
Such a graph construction method ensures fusion of common information elements from utterances within the same topic segment. 
We introduce an \textit{anaphora resolution} step when merging dependency graphs. 
We also introduce an \textit{ambiguity resolver} that takes into consideration the context of words when fusing several utterances. %
Consider the following two utterances:
\begin{itemize}
 \item[] \small{\textit{``um there's a sample sensor and there's a sample speaker unit''}},
 \item[] \small{\textit{``I'm not sure how the sample unit gonna work.''}}
\end{itemize}

Once the first utterance is added into the graph, two nodes containing the word ``\textit{sample}'' are created. 
The ambiguity resolver maps the word ``\textit{sample}'' from the second utterance to the second node (``\textit{sample''} node adjacent to ``\textit{unit}'') 
to account for the correct context of the words. Our goal is to retain the most informative nodes (words) in the graph. 
Further, linguistically well-formed grammatical relations should be retained. 
We formulate the sub-graph generation problem as an Integer Linear Programming (ILP) problem by adapting an existing sentence fusion technique~\cite{filippova2008sentence}.
The solution to the ILP problem generates a sub-graph that satisfies several constraints 
to maximize information content and linguistic quality. Information content is measured using Hori and Furui's word informativeness formula~\cite{hori2003new} while the linguistic quality is estimated using probabilities of grammatical relations from the Reuter's corpus~\cite{apte1994automated}. Grammatical relations extracted from the Reuter's corpus assign higher preferences to non-conversational style of text, thereby resulting in summaries that mirror the flair of human-written abstracts. In the ILP problem, we introduce constraints to limit the length of the sentences. Further, we ensure connectivity in the graph. We introduce several linguistic constraints to generate grammatical output. 
The sub-graph generated from each segment is linearized~\cite{Filippova:2009:TLE:1620853.1620915} using a bottom-up approach to generate a one-sentence summary. 
The one-sentence summaries from all the segments are combined in the final summary. 
Note that we do not introduce any new phrases or words in the process of combining information from multiple utterances. Instead, we consider utterances that are associated with the same topic and apply the ILP-based fusion technique to identify grammatical relations that contains more informative phrases, at the same time leading to generation of summaries that are fairly readable.

To the best of our knowledge, this is the first work that addresses the problems of readability, grammaticality, and content selection jointly for meeting summary generation without employing a template-based approach. 
Experimental results on the aforementioned AMI corpus that consists of meeting recordings show that our approach outperforms the comparable systems. 
ROUGE-2 and ROUGE-SU4~\cite{lin2004rouge} scores from our abstractive model (0.048 and 0.087) are significantly better than that of the extractive summaries (0.026 and 0.044) as well as the word-graph based abstractive summarization method~\cite{mehdad2013abstractive} (0.041 and 0.079). We also assess readability of the summaries using a human judge, demonstrating that the summaries generated by our method are fairly well-formed.  

\section{Related Work}

In the field of meeting summarization, while extractive techniques 
have been widely employed so far~\cite{lin2009graph,liu2010using}, 
abstractive techniques, including sentence compression, template and graph-based approaches, have been focused on recently. 

Liu and Liu~\shortcite{liu2009extractive} used sentence compression to generate summaries of meetings. 
However, they reported that the quality of the generated summaries are not so good and there is a potential limit 
to apply such methods to summarization. 
%
Murray \textit{et al.}~\shortcite{murray2010interpretation} mapped conversations to an ontology that was complemented with a Natural language generation (NLG) component used for transforming utterances to summaries. The corresponding full summarization system was later presented in~\cite{murray2010generating}, where a user study was conducted on the abstractive summaries that were generated. However, the full summarization system involved extensive manual labor to set specific speakers, entities, etc. in a template before using an NLG realizer to generate the summaries. 
Lu and Cardie~\shortcite{wang2013focused} proposed a method that learns templates from the human written summaries and generates the summaries of decisions and actions of meetings by using the best set of templates for a particular summary ranked using a greedy approach. In contrast, we cannot use templates because we assume that the type of a conversation (action, decision, etc) is not known apriori. 

Mehdad~\textit{et al.}~\shortcite{mehdad2013abstractive} developed a method that first over-generates multiple fused utterances in an entailment graph, 
and then chooses one based on the final path ranking. The fusing of the utterances only considers words, and ignores the grammatical relations between them. 
This results in generation of summaries with poor linguistic quality. 
More recently, 
Oya~\textit{et al.}~\cite{W14-4407} used the same fusion technique to generate summaries of meetings. 
Both of the methods developed by Mehdad~\textit{et al.} and Oya~\textit{et al.} mentioned above relied on using multi-sentence compression (MSC)~\cite{filippova2010multi} that combines information from sentences that are similar or connected using some common entity. The MSC technique is a word-graph based method where multiple sentences or utterances can be represented as a network of words. A directed graph is generated where the nodes represent the words while edges exist if two words are adjacent in the utterances. From the graph, several paths between the start and end points can be generated. The new paths can represent content that can be different from the original utterances. 
Oya~\textit{et al.}'s proposed approach requires significant effort to generate the templates using hypernym information for creating slots in the templates. 
Our framework also consists of a similar segmentation module as employed in Oya~\textit{et al.}'s work, which ensures that we divide the meeting transcript into several topics. 
Our proposed method is fundamentally different from most of the aforementioned techniques (except Oya~\textit{et al.}'s work) 
in that it considers individual segments to generate a summary sentence. 

Our previous work~\cite{Banerjee:2015:AMS:2740908.2742751} has briefly described the effectiveness of the fusion-based technique, 
which is also employed in this work. Our preliminary results demonstrated that the fusion based model can combine and convey useful information, 
generating reasonable abstractive meeting summaries. 
Hence, we extend this work using topic segments to build an end-to-end framework. 
We address the issue of readability of the generated summaries by modeling the strength of grammatical relations in the optimization problem. 
Our approach does not require creation of templates. 
Instead, our model aims to generate a sentence on each topic by identifying relevant grammatical relations and informative words from a collection of important utterances in a meeting.

\section{Proposed Approach}
\label{sec:appr}
As explained in Section~\ref{sec:Intro}, 
our proposed approach consists of three steps: First, we segment an entire conversation between participants into multiple text segments. 
Second, we apply an extractive summarizer that extracts important utterances from each segment. 
Finally, we fuse  all the utterances in a segment using an ILP based approach to generate a summary sentence. All the generated sentences are appended to create the final summary. 
In the following, we detail each step. 

\subsection{Text Segmentation}
\label{subsec:TxtSeg}
Topic segmentation has been used in summarization of news articles~\cite{lin2000automated,boguraev2000discourse}. 
Generally, lexical cohesion-based measures work well for topic segmentation~\cite{morris1991lexical}. 
As the primary focus of our work is to generate summaries, 
we experiment with two different text segmentation algorithms: 
\textbf{LCSeg} and \textbf{Bayesian unsupervised topic segmentation}. \\

\begin{table}[t]
 \centering
  \caption{
%
Features to select important utterances. 
Most of them are adopted from previous works~\cite{galley2006skip,xie2008using}. 
The most important speaker refers to the one that utters maximum number of words. 
Our work introduces the segment-based features. The content words include nouns, adjectives, verbs and adverbs.}
		\begin{tabular}{|ll|}
			\hline
			\textbf{(1)} & \textbf{Basic features}\\
			\hline
			 & -- Length of a dialogue \\
			 & -- Number of content words \\
			 & -- Portion of content words \\
			 & -- Number of new nouns introduced \\
			\hline
			\textbf{(2)} & \textbf{Content features}\\
			\hline
			& -- Cosine similarity with entire meeting transcript\\
			& -- Presence of proper nouns\\
			& -- Most important speaker in meeting\\
			& -- Content words in previous dialogue act\\
			\hline
				\textbf{(3)} &  \textbf{Segment based features}\\
			\hline
			& -- Most important speaker in segment\\
			& -- Cosine similarity of dialogue with entire segment\\
			\hline
			\end{tabular}
 \label{tab:extractFeatures}
\end{table}

\noindent
\textbf{LCSeg:} 
Galley \textit{et al.}~\cite{galley2003discourse} developed a topic segmentor, LCSeg, based on lexical cohesion, 
which is considered to be a good indicator of the discourse structure of the text. 
The intuition behind this algorithm is that major term repetitions occur when the underlying topics in the text start or end. 
It takes into consideration multiple features such as discourse cues and overlaps. 
LCSeg is applied to meeting corpora and achieved promising results. 
Hence, this approach is suitable for our segmentation step. \\

\noindent
\textbf{Bayesian unsupervised topic segmentation:}
This is also promising approach to topic segmentation. 
Eisenstein and Barzilay~\shortcite{eisenstein2008bayesian} proposed an unsupervised approach to topic segmentation based on lexical cohesion modeled by a Bayesian framework. 
The cohesion arises through a generative process. The words are modeled from a multinomial language model and the observed likelihood is maximized to generate 
a lexically-cohesive segmentation. 
This algorithm requires a user to specify the desired number of segments. 

\subsection{Selection of Important Utterances}
\label{subsec:UT-Selection}
Our second step is to identify the set of important utterances in each topic segment.
As shown in Table~\ref{tab:extractFeatures}, 
we use multiple features to identify the important set of utterances in a meeting.  
We adopt basic and content features from previous works~\cite{galley2006skip,xie2008using}. 
In addition to the above mentioned features, we introduce two segment-based features:
\begin{itemize}
\item[(i)] The most important speaker in a segment.
\item[(ii)] Cosine similarity between the utterance and all of the other utterances in a segment.
\end{itemize}
We construct classifiers using all the features on the training set. 
We conduct experiments to evaluate the impact of our introduced segment-based features 
in addition to the basic and content features. 
Moreover, constructing a model using the meeting summarization data also suffers from the unbalanced data problem as only few utterances are considered to be important 
to generate the final summary. 
In order to address this problem, we apply the following techniques to oversample the minority data: \\

\noindent{\textbf{Weight:}} Let $npRatio$ be $\frac{\#negative}{\#positive}$. For the training instances, 
we assign weights of one and $npRatio$ to the negative and positive examples, respectively. \\

\noindent{\textbf{Resampling:}} 
We reproduce a random subsample of the training data using sampling with replacement.
In this case, the new training data has the same total number of samples as the old one. 
However, they contain equal populations in both of the classes. \\

\noindent{\textbf{SMOTE:}} In synthetic minority oversampling technique (SMOTE)~\cite{chawla2011smote}, the minority class is randomly oversampled. 
This algorithm forms new examples of minority class by interpolating between several minority class examples that lie together and thereby can avoid the overfitting problem.

%
\subsection{Fusion of Utterances for Summary \\ Generation} \label{subsec: Fusion of Utterances}
The final step in our approach is to combine information from multiple extracted utterances in each segment that the classifier identifies as summary-worthy.  
%
Several techniques have been proposed for sentence fusion tasks~\cite{barzilay2005sentence}. 
However, fusion on meeting utterances requires 
an algorithm that is robust for noisy data as 
utterances often have disfluencies. 
We adapt a sentence fusion technique~\cite{filippova2008sentence} to meeting utterances.
The dependency parse trees of the individual utterances within a topic segment are combined together. The best sub-graph 
that satisfies several constraints and maximizes the propagated information is selected using as an integer linear programming (ILP) formulation.  
%
ILP has been applied successfully to many natural language processing tasks~\cite{clarke2008global,roth2004linear}. 
The formulation of the objective function in the ILP function takes into consideration the informativeness of the words, 
weights of the edges along the dependency tree and a factor that assigns more weights to utterances that are more closer to topic shifts, 
\textit{i.e.}, towards the end of a segment. 
We also introduce an additional step of pronoun resolution. 
We observe that a lot of pronominal references are used in utterances 
and resolving such references would produce more relevant fusion by merging dependency graphs. 
Finally, the solution of the ILP problem is linearized to produce a sentence. 
In this section, we explain all of the details using a simple example. 
Suppose that the following three utterances within a topic segment are labeled as important by the classifier: 
%
%
\begin{itemize}
 \item[($Ut_{1}$)] \textit{``Um well this is the kick-off meeting for our project.''}
 \item[($Ut_{2}$)] \textit{``so we're designing a new remote control and um.''}
 \item[($Ut_{3}$)] \textit{``Um, as you can see it's supposed to be original, trendy and user friendly.''}
\end{itemize}

As can be seen, there are the introductory statements in a meeting that discusses the purpose of the meeting.
We apply pre-processing to get rid of words such as ``\textit{um},'' ``\textit{ah}'' that cause disfluencies 
and do not contribute to any information content in the utterances. \\

\noindent
\textbf{Anaphora resolution.} Our final goal is to generate a one-sentence summary from these utterances. To obtain a summary for each segment, we fuse the dependency graphs of the utterances by merging them on the common words that represent the nodes in the graph. However, in the above example, there are no common words in the three utterances. As can be seen from the utterances, the ``\textit{it}'' in utterance ($Ut_{3}$) refers to ``a new remote control'' in ($Ut_{2}$). To ensure accurate dependency graph merging, where the graphs are merged on the nodes (words in an utterance), it is important to resolve such pronominal references. Without resolving such references, it would be impossible to fuse the above utterances, even though they are referring to the same entity. 
We use the publicly available Stanford CoreNLP package\footnote{\small\url{http://nlp.stanford.edu/software/corenlp.shtml}}~\cite{manning-EtAl:2014:P14-5} 
that has a co-reference resolution module. We resolve pronouns only if there is a pronominal reference to the previous utterance.\\

\noindent\textbf{Dependency graph merging.}
Once anaphora resolution has been applied, the extracted utterances in each segment are parsed using the Stanford dependency parser that is also a part of the CoreNLP package. 
Every individual utterance has an explicit ROOT vertex. We add two dummy nodes in the graph: the \textit{start} node and the \textit{end} node. The ROOT nodes from the utterances are all connected to the \textit{start} node and the last word of every utterance is connected to the \textit{end} node. The words from the utterances are iteratively added onto the graph. 
The words that have the same word form and the parts of speech (POS) tag are assigned to the same nodes. While only content words are merged, stopwords are not merged. 
The use of POS information prevents ungrammatical mappings. 
Hereafter, we refer to a word as the tuple of ``\{word, POS\}.''
We also address ambiguities in the word mappings. 
If a new word that needs to be merged onto the graph has 
multiple mapping candidates, we introduce an ambiguity resolver. \\
\begin{figure*}[t]
 \centering
 \fbox{\includegraphics[scale=0.32]{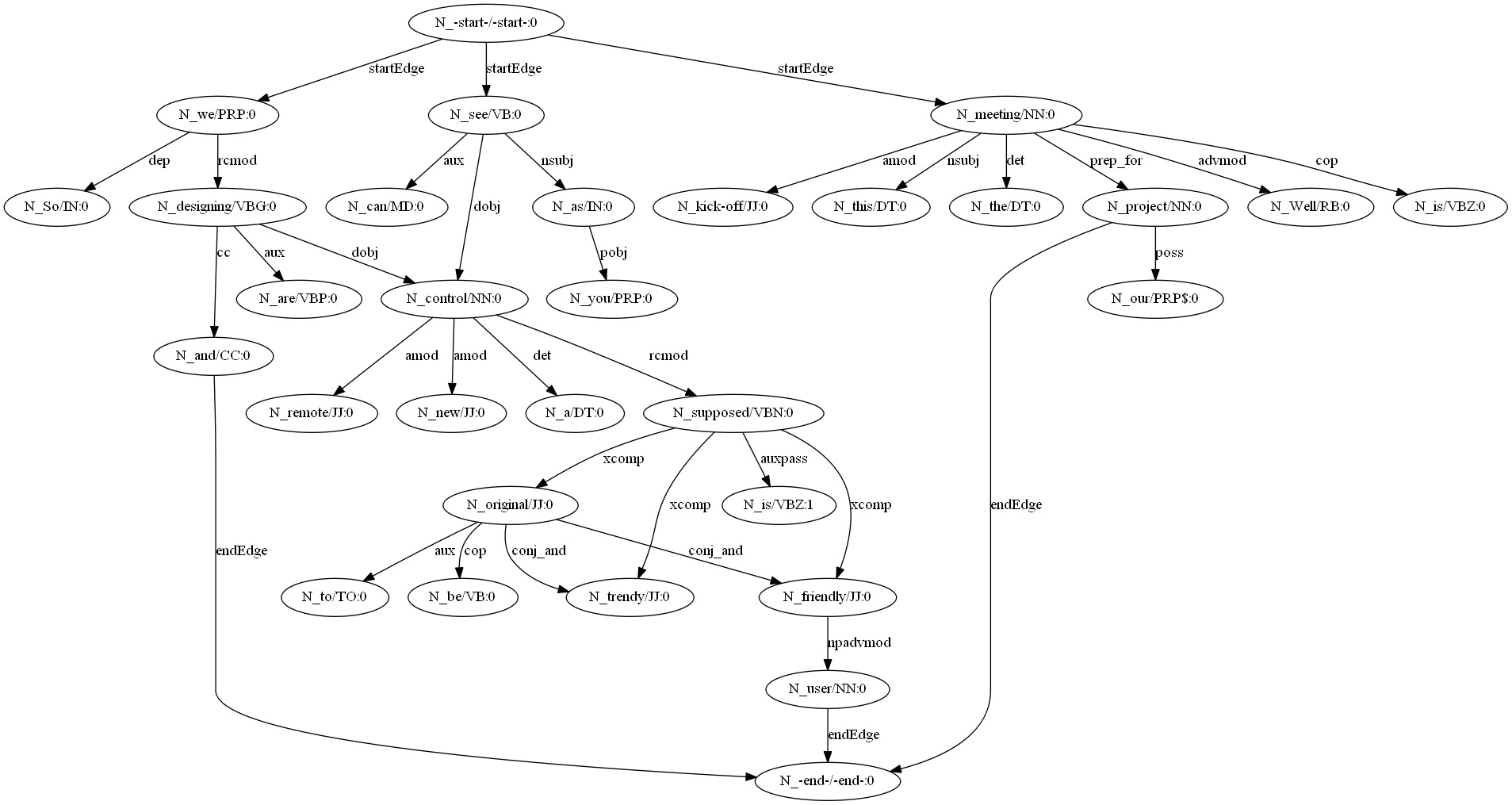}}
 \caption{A merged dependency graph generated from several utterances. Note that only a section of the entire graph is shown. 
The nodes are shown as ``\textit{N\_word}'' and the labels are placed on the edges.}
 \label{fig:dependencyMerge}
\end{figure*}

\noindent
\textbf{Ambiguity resolver.} 
Suppose that a new word $w_{i}$ that has $k$ ambiguous nodes where it can be mapped to. The $k$ ambiguous nodes are referred to as mappable nodes. For every ambiguous mapping candidate, 
we first find the words to the left and right of the mappable word of the sentences, and then compute the number of words in both of the directions that are common to the words in either direction of the word $w_{i}$. We define the directed context as follows: 
\begin{eqnarray}
dirContext = \#CommonWords(dir, window), \nonumber
\end{eqnarray}
where $dir$ and $window$ denote the direction of context (\textit{left}/\textit{right}) and the number of words to be considered in either direction, respectively. 
We calculate the directed context in both of the directions upto a window size of two words. Finally, $w_{i}$ is mapped to the node that has the highest directed context. 
If a tie cannot be broken or no common context can be found with any of the existing nodes, a new node for $w_{i}$ is created. An example of the ambiguity resolution has been provided in Section~\ref{sec:Intro}.

We use the JGrapht\footnote{\small\url{http://jgrapht.org/}} package for the generation of the graph structure. 
Figure~\ref{fig:dependencyMerge} shows a snapshot of the merged dependency graph generated from the three utterances, $Ut_{1}$, $Ut_{2}$, and $Ut_{3}$ 
in Section~\ref{subsec: Fusion of Utterances}. 
The three utterances have been combined together in a common structure, with various possible paths between the \textit{start} and the \textit{end} dummy nodes. To obtain the dependency relations, we use the ``collapsed dependency representation'' from the Stanford parser that collapses edges of conjunctions and prepositions and places the corresponding information on the edge labels (\textit{e.g.}, conj\_and, prep\_at, etc). \\

\noindent
\textbf{ILP formulation.} The next step is to solve and generate a sub-graph from this structure that satisfies a number of syntactic constraints 
and maximizes the information content simultaneously. 
 
Similar to the fusion technique by Fillipova and Strube~\shortcite{filippova2008sentence}, we model the problem as an integer linear programming (ILP) formulation. 
However, the formulation of our objective function and the constraints are significantly different from their system. 
We add a lexical cohesion component in the ILP formulation. Moreover, our constraints leverage linguistic knowledge to generate grammatical output. 
Furthermore, they applied it to German language only. 
The directed edges in the graph are represented as $x_{g,d,l}$ in the ILP problem where $g$, $d$ and $l$ denote the governor node, dependent node 
and the label of an edge, respectively. The edges represent the variables in the objective function which can either take value of 1 or 0 depending on 
whether the edge has to be preserved or deleted.  

We maximize the following objective function: 
\begin{equation}
\sum\nolimits_{x}{x_{g,d,l}\cdot p(l\mid g) \cdot I(d) \cdot \frac{p_{x}}{N} }. 
\label{eqn:ilp}
\end{equation}

As shown in Equation~(\ref{eqn:ilp}), we introduce three different terms: $p(l\mid g)$, $I(d)$ and $\frac{p_{x}}{N}$. 
The term $p(l\mid g)$ denote the probabilities of the labels given a governor node, $g$. We can calculate these probabilities from any given corpora. For every node (word and POS) 
in the entire corpus, the probabilities are represented as the ratio of the sum of the frequency of a particular label and the sum of the frequencies of all the labels emerging from 
a node. In this work, we calculate these values using Reuters corpora~\cite{rose2002reuters} in order to obtain dominant relations from non-conversational style of text. For example, 
Table~\ref{tab:RelationProbs} shows the probabilities of outgoing edges from a node (``\textit{produced/VBN}''). 
\begin{table}[t]
 \centering
 \scriptsize
 \caption{Probabilities of outgoing edges from a node for ``\textit{produced/VBN}.''} \label{tab:RelationProbs}
 \begin{tabular*}{0.47\textwidth}{c|c|c|c|c|c|c}\hline
   auxpass&nsubjpass&aux&prep\_with&agent&prep\_in&advmod\\ \hline
   0.286&0.214&0.214&0.071&0.071&0.071&0.071\\ \hline	
 \end{tabular*}
\end{table}
The term $I(d)$ denotes the informativeness of a node. 
In order to compute $I(d)$, 
we improve the word significance score~\cite{hori2003new} as follows:
\begin{equation}
 I(d)=f_{s}\cdot\log\frac{F_{A}}{F_{d}}, \label{eq:Informativeness}
\end{equation}
where $f_{s}$, $F_{A}$, and $F_{d}$ denote the frequency of a word in a text segment, the sum of the frequencies of all the words in the corpus, and the frequency of the dependent word $d$ in the entire Reuters corpus, respectively. 
The last term $\frac{p_{x}}{N}$ in Equation~(\ref{eqn:ilp}) is based on the idea of lexical cohesion. 
Our intuition is that important decisions in a meeting are taken just before a topic concludes. 
Therefore, to model the relative importance of such utterances, we introduce the term $\frac{p_{x}}{N}$, 
where $N$ and $p_{x}$ denote the total number of extracted utterances in a segment and the position of the utterance (the edge $x$ belongs to) in the set of $N$ utterances, respectively. As a result of this term, utterances more closer to topic boundaries are assigned higher weights. 

In order to solve the above ILP problem, we impose a number of constraints. 
Some of the constraints have been directly adapted from the original ILP formulation~\cite{filippova2008sentence}. 
For example, we use the same constraints for restricting one incoming edge per node, as well as we impose the connectivity constraint to ensure a connected graph structure. The other constraints we impose are defined as follows:
\begin{equation}
\begin{aligned}
\forall{l \in startEdge},\sum\nolimits_{l}{x_{g,d,l}}=1, \\
\forall{l \in endEdge},\sum\nolimits_{l}{x_{g,d,l}}=1
\end{aligned}
\label{eqn:startend}
\end{equation}  
\begin{equation}
\sum\nolimits_{x}{x_{g,d,l}}\leq \gamma
\label{eqn:length}
\end{equation}
\begin{equation}
\sum\nolimits_{g,d}{( x_{g,d,l}+x_{d,g,l} )} \leq 1
\label{eqn:nocycle}
\end{equation}
\begin{equation}
\forall{l_{out} \in \{aux,cop,det\}}, \sum\nolimits_{u,l_{in}}{x_{g,u,l_{in}}}-x_{u,d,l_{out}}=0
\label{eqn:auxcop}
\end{equation}
\begin{equation}
\forall{g, l_{out} \in aux\lor cop \lor det}, \sum\nolimits_{l_{out}}{x_{g,d,l_{out}}} \leq 1
\label{eqn:det}
\end{equation}

Equation~(\ref{eqn:startend}) limits the subtree to compulsorily have just one start edge and one end edge. This helps in preserving one ROOT node, as well as it limits to one \textit{end} node for the generated subtree. Equation~(\ref{eqn:length}) limits the generated subtree to have a maximum of $\gamma$ nodes. The start nodes and end nodes are still a part of the subtree that is generated by solving this optimization problem. Hence, the value of $\gamma$ needs to be set to 2, which is more than the desired number of maximum words in the summary sentence. 
In order to prevent bidirectional relations between two nodes, we impose Equation~(\ref{eqn:nocycle}) as a constraint. 
To maintain the linguistic quality of the generated sentence, by using Equations~(\ref{eqn:auxcop}) and~(\ref{eqn:det}) as constraints, 
we always include a maximum of one auxiliary verb (aux), copular verb (cop) and determinant (det) if they exist. 
We use the Gurobi software\footnote{\small{\url{http://www.gurobi.com/}}}~\cite{gurobi} for the optimization tasks.
\begin{figure}[t]
 \centering
  \fbox{\includegraphics[width=0.42\textwidth]{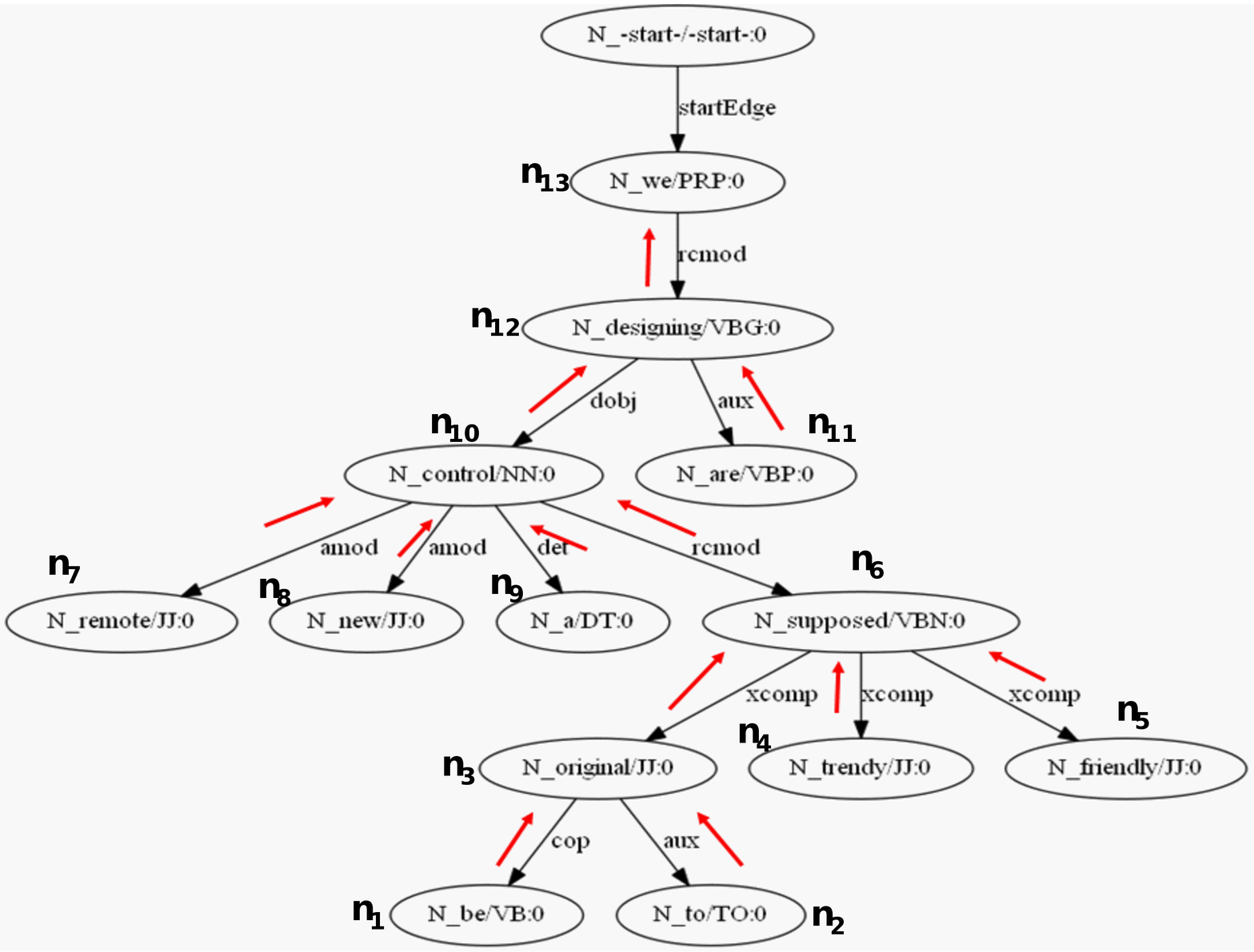}}
  \caption{Dependency graph linearization process.}
  \label{fig:FinalDependencygraph}
\end{figure}
Figure~\ref{fig:FinalDependencygraph} shows the final graph that is retained from the graph in Figure~\ref{fig:dependencyMerge} after solving the ILP problem. \\

\noindent
\textbf{Linearization:}
The purpose of linearization is to generate a sentence from the final subtree generated by solving the ILP problem. We take a relatively straightforward bottom-up approach to tackle the problem. We keep adding the leaf nodes to their governing nodes until it reaches the ROOT node. We maintain the same order of the words as in the source sentences during the merging process.

As shown in Figure~\ref{fig:FinalDependencygraph}, the nodes \textit{to/TO ($n_{2}$)} and \textit{be/VB ($n_{1}$)} are added to the node \textit{original/JJ ($n_{3}$)}. 
After merging these words, they are reordered so that the ordering resembles the one in the source sentences. 
The sequence of the nodes is changed only during the merge with the governing node: they are kept fixed for the future operations. 
Hence, the nodes \textit{to/TO ($n_{2}$)} and \textit{be/VB ($n_{1}$)} are memorized along with the node \textit{original/JJ ($n_{3}$)}. 
In the next step, the ordering of the node \textit{original/JJ ($n_{3}$)} matters with respect to the other leaves the governing node ($n_{6}$) has. 
However, this might be a problem in the case where there are leaf nodes from a governor node at some higher level. 
In this example, \textit{trendy/JJ ($n_{4}$)} and \textit{friendly/JJ ($n_{5}$)} will get merged to \textit{supposed/VBN ($n_{6}$)} before the node \textit{original/JJ ($n_{3}$)} as they are leaf nodes. 
To prevent such merging, we only allow to merge the leaf nodes to the governing node that is at the farthest distance from the ROOT vertex. We apply Dijkstra's algorithm~\cite{skiena1990dijkstra} to calculate the path length. 
Thus, in Figure~\ref{fig:FinalDependencygraph}, the nodes \textit{trendy/JJ ($n_{4}$)} and \textit{friendly/JJ ($n_{5}$)} are added only after \textit{to/TO ($n_{2}$)} and \textit{be/VB ($n_{1}$)} are merged to \textit{original/JJ}. The final sentence after linearization is as follows:
\begin{itemize}
\item[] \textit{We are designing a new remote control supposed to be original trendy and friendly. }
\end{itemize}

\section{Experimental Results}
\subsection{Dataset and Evaluation Metrics}
The AMI Meeting corpus~\cite{carletta2006ami} contains 139 meeting transcripts along with their corresponding extractive and abstractive summaries. 
The standard test set of this corpus includes 20 meetings. 
Our extractive summarization component is trained using the training set, \textit{i.e.}, the remaining $119$ meetings. 
We evaluate the accuracy of the classifiers using standard metrics: Precision, Recall and F-measure. 
We also evaluate the impact of introducing segment-based features. 
To evaluate the quality of the summaries, 
we verify the effectiveness of content selection using ROUGE, which has been widely used as a standard technique to evaluate information content 
in document summarization tasks by comparing system-generated summaries to human-written abstractive summaries. 
Further, we also evaluate the linguistic quality of the generated summaries using human judgments.
\begin{figure}[t]
 \centering
 \fbox{\includegraphics[width=0.41\textwidth]{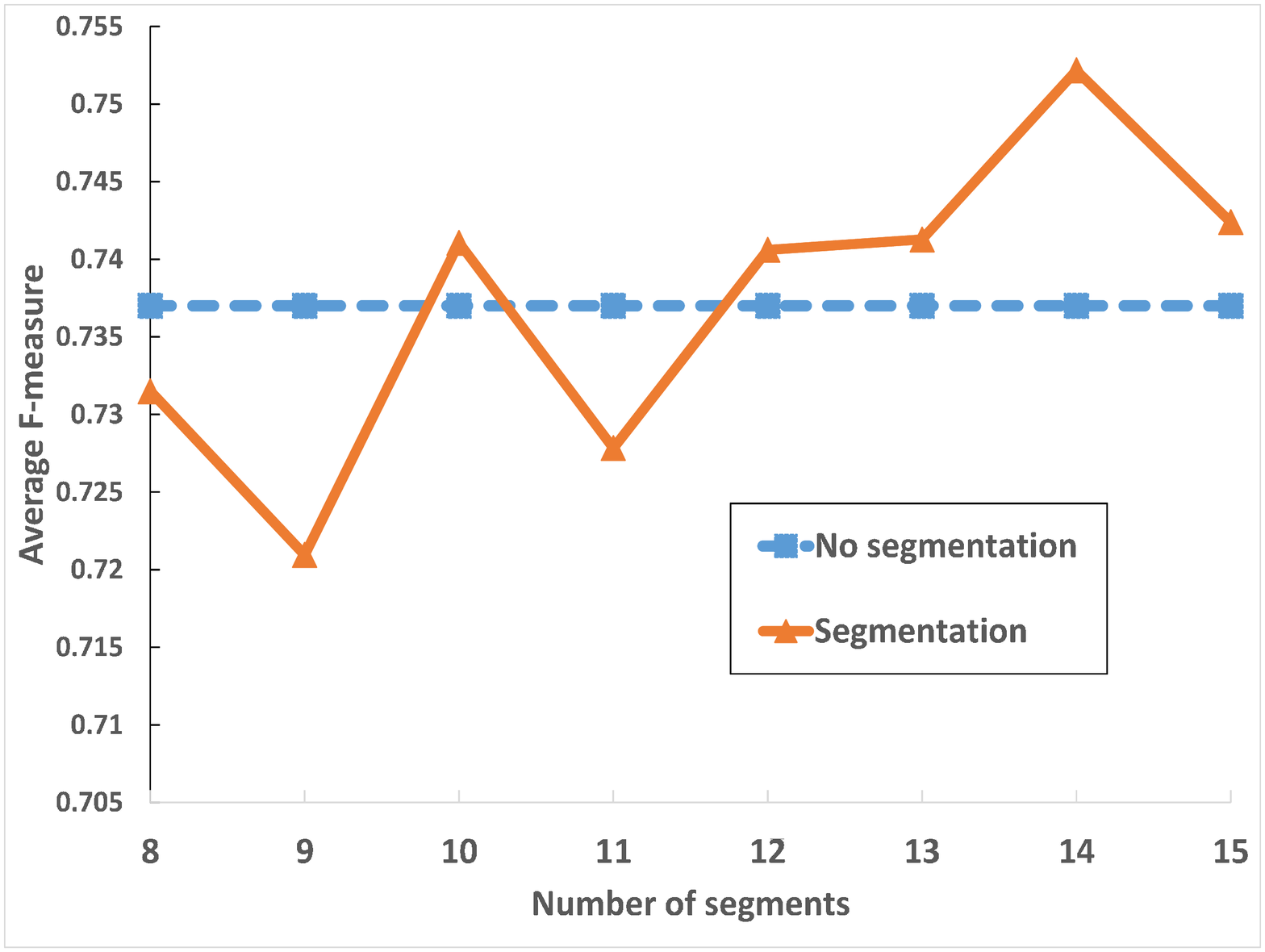}}
 \caption{Average F-measures obtained by varying the number of segments. Also shows the impact of addition of segment-based features.}
 \label{fig:segnonseg}
\end{figure}

\vspace{-2mm}
\subsection{Classifier Evaluation}
As described in Section~\ref{subsec:TxtSeg}, 
we used two different text segmentation algorithms: LCSeg and Bayesian unsupervised topic segmentation. 
Furthermore, we employed three classifiers: Support Vector Machines (SVM), Random Forest (RF) and Naive Bayes (NB). 
To overcome the problem of unbalanced data, we used three different sampling techniques 
(Weight, Resampling, and SMOTE) as described in Section~\ref{subsec:UT-Selection}. 
We evaluated all the possible configurations on the training dataset to determine the best configuration suitable for 
our summary generation. We used Weka~\cite{hall2009weka} for all the classification tasks with the default set of parameters. 
We perform 10-fold cross validation on the training set. 
First, we try to find the optimal number of segments that provides the best classification accuracy. 
Simultaneously, we also evaluate the contribution of adding segment-based features during training the classifiers. 
Second, we also identify the text segmentation algorithm that works best on this dataset. 
Finally, based on the above decisions, we compare the performances of the classifiers to decide our extractive summarization component 
(\textit{i.e.}, classifier that gives the the best) 
and the best sampling strategy to avoid any bias due to the unbalanced dataset. \\

\vspace{-2mm} 
\noindent
\textbf{Number of segments:}
We optimize the number of segments for each meeting by varying it from 8 to 15 on the training data. 
Figure~\ref{fig:segnonseg} shows the average F-measure obtained by the classifiers constructed from the set of all features (``Segmentation'') 
and the features excluding the segment-based features (``No segmentation''). The graph shows the average F-measure obtained by all combinations of 
the classifiers and the sampling strategies with respect to the various number of segments. 
As expected, the average F-measures do not show any change when we skip the segment-based features. 
However, we observe slight differences when we introduce the segment-based features. 
Generally, when the number of segments is between 12 to 15, 
we observe about 1\% improvement in F-measure by adding the segment-based features over the basic set of features. 
According to Figure~\ref{fig:segnonseg}, 
we observe the highest average F-measure of 0.752 when we segment the meeting transcript into 14 topics. We set the number of segments to 14 for our following experiments. \\
\begin{figure}[t]
 \centering
 \fbox{\includegraphics[width=0.41\textwidth]{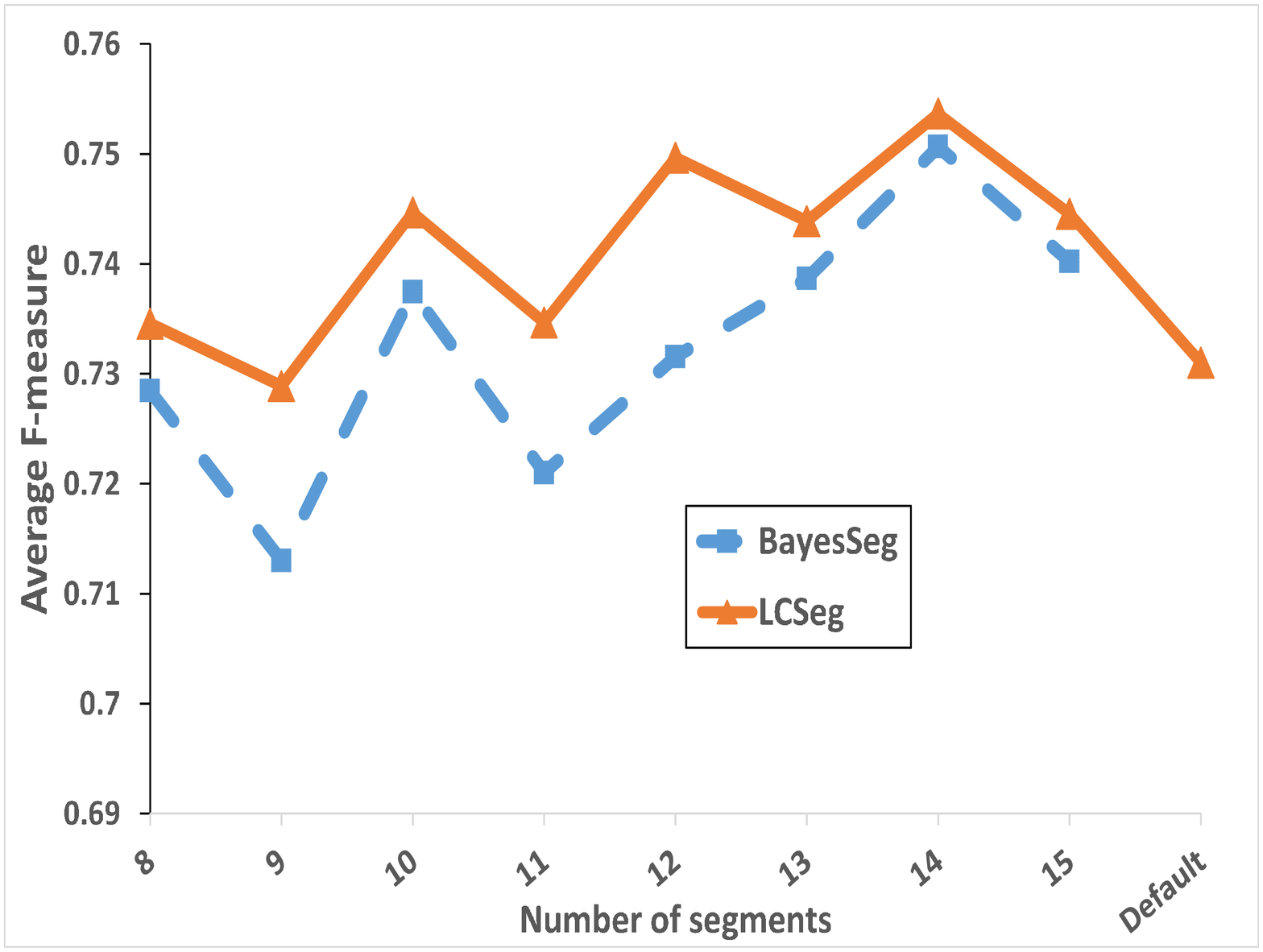}}
  \caption{Comparison of performance obtained by text segmentation algorithms: Bayesian segmentation (BayesSeg) and LCSeg. ``Default'' setting does not require the number of segments to be explicitly stated when computing segment boundaries using LCSeg.}
 \label{fig:bayeslc}
\end{figure}

\noindent
\textbf{LCSeg vs Bayesian unsupervised topic segmentation:}
Figure~\ref{fig:bayeslc} shows the comparison of average F-measures of the classification models for each of the topic segmentation algorithms. As can be seen, LCSeg generally outperforms Bayesian segmentation. The \textit{default} setting of the number of segments refers to the setting for LCSeg,  
which does not require to specifying the number of text segments. 
According to Figure~\ref{fig:bayeslc}, the default setting does not perform well. 
Both of the segmentation algorithms achieve similar F-measure in classification accuracies when we set the number of segments to 14. Note that the optimal number of segments were obtained by applying 10-fold cross validation using both the segmentation algorithms. \\
\begin{table}[t]
\scriptsize
\caption{Precision (Pre), Recall (Rec), and F-measure (F) obtained by classifiers along with each of the three sampling strategies. We set the number of segments to 14.}
\begin{tabular}{ c | l | c | c | c | c |c |c}
\hline
\multicolumn{2}{c|}{}  & \multicolumn{3}{c|}{\textbf{LCSeg}} & \multicolumn{3}{c}{\textbf{BayesSeg}}  \\ \hline
	\textbf{Classifier} & \textbf{Sampling} & \textbf{Pre} & \textbf{Rec} & \textbf{F} & \textbf{Pre} & \textbf{Rec} & \textbf{F} \\ \hline
	&Weight& 0.792	&0.787&	0.786&	0.788&	0.784	&0.783\\ 
	NB &Resampling  & 0.796 &	0.790&	0.789&	0.791	&0.787&	0.786\\ 
	& SMOTE & 0.831 &	0.804	&0.811&	0.830&	0.807&0.813\\ 
\hline
	&Weight & 0.726&0.722&	0.721&	0.730&	0.727&	0.727\\
	RF &Resampling  & 0.892	&0.889 &	\textbf{0.888}&	0.880&	0.877&\textbf{0.877}\\
	& SMOTE  & 0.817 &0.821&0.819&0.817&0.820&0.818\\
\hline
	&Weight  & 0.623&	0.591&	0.563&	0.601&	0.584&	0.566\\
	SVM &Resampling  & 0.752	&0.741	&0.739&	0.690&	0.689&	0.689\\
	& SMOTE  & 0.660	&0.694&	0.667&	0.691	&0.711	&0.697\\
	\hline
\end{tabular}
\label{tab:classificationAccuracy}
\end{table}

\noindent
\textbf{Evaluation of classifiers and sampling strategy:} Table~\ref{tab:classificationAccuracy} shows the results of classification evaluation. 
The scores in the table were obtained by setting the number of segments to 14. As can be seen from the table, 
the NB and RF classifiers significantly outperform SVM. 
The best system is obtained by combining the RF classifier with the Resampling strategy. 
In both of the segmentation algorithms, the combination of RF and Resampling gives the best F-measure (0.888 and 0.877). 
Resampling and SMOTE sampling strategies outperform the Weight strategy when using RF and SVM. However, when using NB, 
the performance using Weight and Resampling strategy is very similar. 

\subsection{Content Selection}
In text summarization, it is also important to evaluate to what extent a classifier retains valuable information that should exist in a summary. 
Therefore, system generated summaries should be compared to human-written summaries automatically.  
We evaluate content selection using ROUGE. \\

\noindent
\textbf{Training Set:} Table~\ref{tab:contentSel} shows the experimental results of content selection on the training set. 
We compare extractive summaries with human-generated abstracts for all the meetings in the training set and compute ROUGE-1 (R-1) and ROUGE-2 (R-2) scores. 
We do not impose any length constraints during ROUGE evaluation on the training data. 
Similar to Table~\ref{tab:classificationAccuracy}, the combination of RF with Resampling strategy outperforms other techniques in terms of R-1 and R-2 scores.  
We use the RF classifier trained using Resampling strategy as our extractive component. This classifier is used on the test set to identify important utterances in the meeting transcripts.
We segment each meeting transcript into 14 (optimal) segments using the LCSeg algorithm 
as it slightly outperform Bayesian unsupervised topic segmentation. \\

\begin{table}[t]
\scriptsize
\centering
\caption{ROUGE scores obtained by several configurations in content selection. We compute ROUGE-1 (R-1) and ROUGE-2 (R-2) without any limit on summary length for comparison.}
\begin{tabular}{ c | l | c | c | c | c }
\hline
\multicolumn{2}{c|}{}  & \multicolumn{2}{c|}{\textbf{LCSeg}} & \multicolumn{2}{c}{\textbf{BayesSeg}}  \\ \hline
	\textbf{Classifier} & \textbf{Sampling} & \textbf{R-1} & \textbf{R-2} & \textbf{R-1} & \textbf{R-2} \\ \hline
	&Weight & 0.660 & 0.141 & 0.663 & 0.142 \\
	NB &Resampling  & 0.666 & 0.142& 0.674 & 0.145 \\
	& SMOTE  & 0.673 & 0.145 & 0.675 & 0.147\\
\hline 
	&Weight&0.679 & 0.147 & 0.702 & 0.144  \\ 
	RF &Resampling  & \textbf{0.705} & \textbf{0.158}& \textbf{0.703} & \textbf{0.157}  \\ 
	& SMOTE &0.694 & 0.152   & 0.695 & 0.153 \\ 
	\hline
	&Weight  & 0.490 & 0.114& 0.473 & 0.112\\
	SVM &Resampling  & 0.563 & 0.143& 0.556 & 0.139 \\
	& SMOTE & 0.525 & 0.123& 0.567 & 0.141\\
	\hline
\end{tabular}
\label{tab:contentSel}
\end{table}

\begin{table}[t]
	\centering
		\caption{Content selection evaluation. 
		We compute ROUGE-2 (R-2) and ROUGE-SU4 (R-SU4) scores by comparing the system generated summaries and the human-written summaries for all the meetings in the test set.}
		\begin{tabular}{l|c|c}
		\hline
		\textbf{Method} & \textbf{R-2} & \textbf{R-SU4}\\
		\hline
		Our abstractive model& \textbf{0.048} & \textbf{0.087} \\
		Our abstractive model \scriptsize{(no anaphora resolution)} & 0.036 & 0.071 \\
		MSC model~\cite{filippova2010multi} & 0.041 & 0.079 \\	
		Extractive model \scriptsize{(baseline)} & 0.026 & 0.044 \\
		\hline
\end{tabular}
	\label{tab:absContentSel}
\end{table}

\noindent
\textbf{Test-set evaluation:} We generate abstractive summaries from the meeting transcripts in the test set using our ILP-based approach. 
To evaluate our abstractive summaries, we compare them to the extractive summaries generated by the best performing classifier. 
We also compare the summaries to the MSC method proposed by Fillipova~\shortcite{filippova2010multi} that has been adapted for 
abstractive meeting summarization~\cite{mehdad2013abstractive} develped earlier.
As an input to the MSC model, we use the same set of utterances per segment that was extracted by the classifier. 
The sentence in each segment that obtains the highest score using MSC is used in the final generated summary. 
The human-written abstracts, on average, contain close to 300 words. 
Therefore, we applied a length constraint while performing ROUGE evaluation to limit summary comparison upto 300 words. 

We use ROUGE-2 (R-2) and ROUGE-SU4 (R-SU4) recall scores to compare all approaches. Both the ROUGE scores have been found to correlate well with 
human judgments~\cite{nenkova2012survey}.
Table~\ref{tab:absContentSel} shows that our abstractive model can effectively maximize information content, resulting in better summaries compared with the other models. 
Furthermore, the ROUGE scores of the MSC model also significantly outperforms the extractive model, indicating that MSC results in more informative summaries. 
To evaluate the impact of anaphora resolution, we run our abstractive summarization model without performing the pre-processing step of pronoun resolution.
ROUGE-2 score obtained by the abstractive model with anaphora resolution (0.048) is significantly better than the model without anaphora resolution (0.036). 
This indicates that anaphora resolution 
significantly contributes to content selection. 
Due to pronoun resolution, there are more chances of fusing information from various utterances within a topic segment.
The extractive model obtains the lowest ROUGE scores as we restrict comparison to the first 300 words. 
In contrast, the abstractive methods (our methods and MSC) can effectively integrate the information from multiple utterances within the first 300 words. \\
\begin{table}[t]
 \centering
 \caption{Readability estimates of summaries.}
 \begin{tabular}{l|c|c}
  \hline
  \textbf{Method}&\textbf{Readability score}&\textbf{Log likelihood}\\ \hline
	Our abstractive model & 0.74 & -\textbf{125.73}\\
        MSC model~\cite{filippova2010multi} & 0.62 & -141.31\\
        Extractive model & 0.67 & -136.22\\\hline
	\end{tabular}
	\label{tab:ReadabilityEstimates}
\end{table}

\noindent \textbf{ROUGE comparison:} In general, shorter summaries are preferred by human readers. Extractive meeting summaries tend to be very long. In contrast, human-written summaries are very short and contain 300 words on average. To take this preference of shorter summaries into account, we evaluate summaries using only the first 300 words. The extractive summaries are fairly long and contain 2000-5000 words. Including several utterances distracts a reader's focus on the salient aspects, resulting in low readability of the summaries.
Therefore, we set the length parameter in ROUGE (l) to 300 to limit the comparison to only the first 300 words. 
\vspace{-1mm}

\subsection{Readability Analysis}
To evaluate the linguistic quality of the generated summaries, we perform readability analysis. 
We asked one human judge to mark sentences in the generated abstractive summaries as either readable or not readable. 
Readability indicates how well the idea in the sentence is conveyed to the reader. 
Excessive presence of disfluencies or ill-formed utterances should be marked as not readable. We provided these instructions to the human judge. 
Out of 261 summary sentences generated for the 20 abstractive summaries, 67 sentences were found to be \textit{not readable} ($\sim$26\%). 
We also manually evaluated the extracted utterances and found that 33\% of the utterances contained various disfluencies, making them difficult to read. 
We also performed another readability analysis for the summaries generated using MSC and found that only 62\% of the generated sentences in the summaries are readable. 
The readability of MSC summaries (0.62) is even worse than that of the extractive summaries (0.67), showing that, 
while the generated sentences in MSC model are informative (high ROUGE scores), 
they suffer from serious grammatical issues as no factor of linguistic quality is considered in the model. 
Generally, several utterances in extractive summaries were marked as not readable due to excessive use of disfluencies. Furthermore, there were incomplete utterances that created confusion in the minds of the reader. For example, an extracted utterances -- ``Ah eagle , right okay .'' -- although grammatical, does not tell us anything about what is being spoken about. 
To obtain a coarse estimate of grammaticality, we also calculate the average log-likelihood score provided by the Stanford Parser. 
We compute the average log-likelihood scores of the confidence of the dependency parses for each type of summaries. 
Table~\ref{tab:ReadabilityEstimates} shows the average scores.\footnote{Lower the magnitude of the log-likelihood scores, 
the higher is the confidence associated with the dependency parse.}  

\begin{table}[t]
	\centering
		\caption{Examples of summary sentences that our system generated \textbf{(S)} and corresponding human--written \textbf{(H)} summary sentences. 
Note that they are only small portions of the summaries and not the entire summaries.}
\begin{tabular}{|>{\everypar{\hangindent 0.4cm}}p{0.45\textwidth}|}
		\hline
		\textbf{S:} Slightly curved around the sides like up to the main display as well.
		It was voice activated .\\
		\textbf{H:} The remote will be single-curved with a cherry design on top. A sample sensor was included to add speech recognition.\\
		\hline
		\textbf{S:} The market trends and our traditional usual market research study suggests the use of rechargeable batteries.\\
		\textbf{H:} He suggested substituting a kinetic battery for the rechargeable batteries and using a combination of rubber and plastic for the materials. \\
		\hline
		\textbf{S:} And in this detailed design the usability interface meeting we will discuss our final design the look-and-feel.\\
		\textbf{H:} All these components were re-arranged in a revised prototype. \\
		\hline
		\end{tabular}

	\label{tab:ExamplesOfSummaries}
\end{table}

Table~\ref{tab:ExamplesOfSummaries} shows some examples of summaries generated by our system. 
The table indicates that the summaries generated by our system are relatively well-formed and they reflect the formal style of non-conversational text. 
We aligned sentences from the human-written abstracts and the corresponding sentences selected from each segment that our algorithm generates. \\

\noindent \textbf{Error Analysis:} In some cases the linearization component does not produce relevant ordering of words. 
For example, in the third summary sentence in Table~\ref{tab:ExamplesOfSummaries}, the system-generated summary lacks certain conjunctions and the ordering of words is inappropriate. The phrase ``detailed design the'' can be removed to maintain clarity. 
In addition, the entity -- \textit{design} has been repeated. 
To solve these problems, we plan to introduce intra-sentence level constraints to improve generated summaries. 

Our algorithm is designed to retain informative words as well as grammatical dependencies that are more probable in any given corpus. However, the grammatical dependencies that we choose might not necessarily lead to well-formed grammatical sentences. Furthermore, our ILP-based model is unable to understand long-term dependencies of entities within a generated sentence. 
For example, consider the following output: \\

\noindent
\textit{``Decided important reflect our budget our the product accessible a wide range of consumers limiting anyone know that kind .''} \\
\vspace{-2mm}

As can be seen, it is really hard for a reader to identify what the summary sentence is trying to convey although certain words or phrases hint at the topic of \textit{deciding the budget based on the range of consumers}. Our model at present does not memorize previous choices of entities referred in the utterances. Furthermore, our linearization component is based on the ordering of words in the source utterances. However, using the same ordering as the source sentences might not necessarily work well. In the context of an entirely new generated summary sentence, lexical or phrasal reordering and other transformations might be required. In such cases, it might be more effective to use language model based confidence scores to determine the best ordering of words. Improvement in the ILP formulation is possible by including confidence of the sequence of words in addition to the incorporation of knowledge about the entities. Optimizing such a complete model can help generate summaries that are much easier to read. Further, such summaries would contain coherent elements on the same entities in the summary sentences. We might also hope to optimize the model by including the number of segments as a parameter in the model. Currently, the maximum number of sentences in the summary is dependent on the number of topic segments. We can improve the formulation such that model itself decides the optimal length of the summaries ensuring that all the informative points in the meeting discussion are included in the system generated summary.

\vspace{-2mm}

\section{Conclusions and Future Work}
In this work, we have proposed an approach to generate abstractive summaries from meeting conversations. We proposed a method for dividing a conversation into multiple topic segments. 
We used an extractive summarizer to identify the important set of utterances, and then applied ILP-based utterance fusion to generate one sentence summary from every topic segment. 
We leveraged the grammatical relations in the fusion technique that are more dominant in non-conversational style of text. 
The experiments on content selection and readability indicate that 
our method can generate relevant abstractive summaries from meeting transcripts without any templates. However, as we have already pointed out, not all generated summaries are usable due to the lack of coherence among several entities discussed within the same summary sentence. We plan to improve the generation using knowledge of entities and also refine readability using a language model.
In future work, we plan to develop better linearization techniques. 
We also plan to improve our algorithm by not limiting one sentence per segment but allowing the ILP model to decide the optimal number of sentences for a complete summary.

\vspace{-2mm}
\section{Acknowledgments}
\vspace{-1mm}
\noindent This material is based upon work supported by the National Science Foundation under Grant No. 0845487.

\vspace{-2mm}
\bibliographystyle{abbrv}
\bibliography{sig-alternate}

\end{document}